\documentclass[11pt,a4paper]{article}
\usepackage[hyperref]{acl2019}
\usepackage{times}
\usepackage{latexsym}

\usepackage{url}
\usepackage{amsmath}
\usepackage{amssymb}
\usepackage{amsfonts}
\usepackage{graphicx}
\graphicspath{{Figure/}}

\aclfinalcopy % Uncomment this line for the final submission
\def\aclpaperid{2274} %  Enter the acl Paper ID here

\title{Know More about Each Other: Evolving Dialogue Strategy via Compound Assessment}

\author{Siqi Bao, Huang He, Fan Wang, Rongzhong Lian \and Hua Wu \\
  Baidu Inc., China \\
  \texttt{\{baosiqi, hehuang, wangfan04, lianrongzhong, wu\_hua\}@baidu.com} \\
  }

\date{}

\begin{document}
\maketitle
\begin{abstract}
In this paper, a novel Generation-Evaluation framework is developed for multi-turn conversations with the objective of letting both participants know more about each other. For the sake of rational knowledge utilization and coherent conversation flow, a \underline{dialogue strategy} which \underline{controls knowledge selection} is instantiated and continuously adapted via reinforcement learning. Under the deployed strategy, knowledge grounded conversations are conducted with two dialogue agents. The generated dialogues are comprehensively evaluated on aspects like informativeness and coherence, which are aligned with our objective and human instinct. These assessments are integrated as a compound reward to guide the evolution of dialogue strategy via policy gradient. Comprehensive experiments have been carried out on the publicly available dataset, demonstrating that the proposed method outperforms the other state-of-the-art approaches significantly.
\end{abstract}

\section{Introduction}
Intelligent dialogue systems have become popular in our daily life, such as the chit-chat XiaoIce and the task-oriented Echo. These systems serve as smart agents to facilitate more effective interaction with users in various situations, like ticket booking or recreation offering. Primary dialogue systems \cite{vinyals2015neural,shang2015neural}  try to mimic human beings to generate fluent utterances, whereas paying little attention to the intrinsic factors of human conversations: \textit{exchanging information} and \textit{enhancing interaction} \cite{li2017dailydialog}. Therefore, they are prone to generate dull and generic responses.

To address this problem, in recent years, several approaches have been developed to generate informative responses based on external knowledge. Recently, a knowledge grounded model is proposed in \newcite{ghazvininejad2018knowledge}, where relevant factual texts are encoded into memory and replies are decoded via attention mechanism. Instead of using unstructured text knowledge, CCM \cite{zhou2018commonsense} relies on structured knowledge to generate rich-information response. However, all these approaches are designed for the single-round settings. While applied to the real-world scenarios (where dialogues are conducted for multiple rounds), the dialogue quality will be severely limited due to the lack of coordination among different rounds.

\begin{figure*}[ht]
	\centering
	\includegraphics[width=\textwidth]{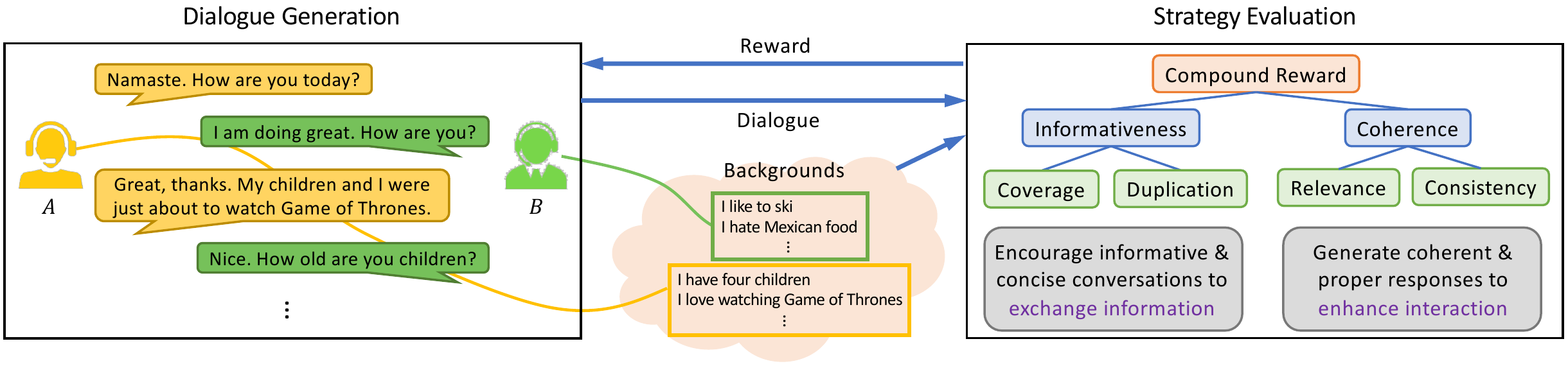}
	\caption{Framework overview. Left: dialogue generation. Right: strategy evaluation.}
	\label{fig:framework}
\end{figure*}
As discussed above, one of the ultimate goals in human conversation is that information can be exchanged effectively through interaction. Particularly, we argue that successful multi-turn dialogues are determined by the joint experience of both participants in the conversation, i.e., both participants need to get aware of their counterparts and express themselves effectively. To this end, we propose the objective of letting both sides know more about each other. With this objective, a novel Generation-Evaluation framework is introduced for the multi-turn dialogues.

As the name Generation-Evaluation indicates, there are two fundamental modules in our framework. In the module of dialogue generation, a two-stage generative model is employed, where the dialogue strategy determines which knowledge to use for the current turn and the decoder uses this knowledge to produce the response. In the module of evaluation, the generated dialogues are assessed from the following two aspects: \textit{informativeness}, which measures the effectiveness of information exchange and \textit{coherence}, which reflects the response's suitableness. Both modules are assembled within a unified reinforcement learning pipeline. The generation module simulates knowledge grounded conversations with two dialogue agents and receives compound reward from the evaluation module. By keeping adapted for higher evaluation rewards, the generation module will be continuously evolving for better dialogue quality. As suggested in \newcite{Yarats2018HierarchicalTG}, applying reinforcement learning on the decoder might bring in adverse impacts on the linguistic quality. As such, in the generation module, the decoder is pre-trained with supervised learning and the dialogue strategy keeps evolving with reinforcement learning.  

The contributions of this work are summarized as follows:
\begin{itemize}
\item With the objective of letting both participants know more about each other, we propose a novel Generation-Evaluation framework, which facilitates the generation of informative and coherent dialogues. 

\item To evaluate the effectiveness of dialogue strategy, two metrics are specially designed on informativeness and coherence, which are further integrated as a compound reward. Towards maximizing this reward, the strategy of knowledge selection is able to evolve via reinforcement learning. 

\item Intensive and extensive experiments have been carried out on PersonaChat. As compared with other state-of-the-art approaches, our method obtains superior performances on both automatic and human evaluations.
\end{itemize}

\section{Methodology}
\subsection{Framework Overview}
Our Generation-Evaluation framework is illustrated in Figure \ref{fig:framework}. Under the deployed strategy of knowledge selection, two dialogue agents introduce themselves alternately in accordance with corresponding backgrounds and make responses to their counterparts in a proper way. The generated dialogues together with the agents' backgrounds are collected for strategy evaluation in terms of two essential aspects: informativeness and coherence. Then these assessments are integrated as a compound reward, acting as the reinforcing signal for the evolution of knowledge interaction strategy.

In the following parts, we will first introduce the process of dialogue generation, present the metrics utilized in strategy evaluation and then describe the strategy evolution via compound assessment. 

\subsection{Dialogue Generation}
\begin{figure}
	\centering
	\includegraphics[width=0.48\textwidth]{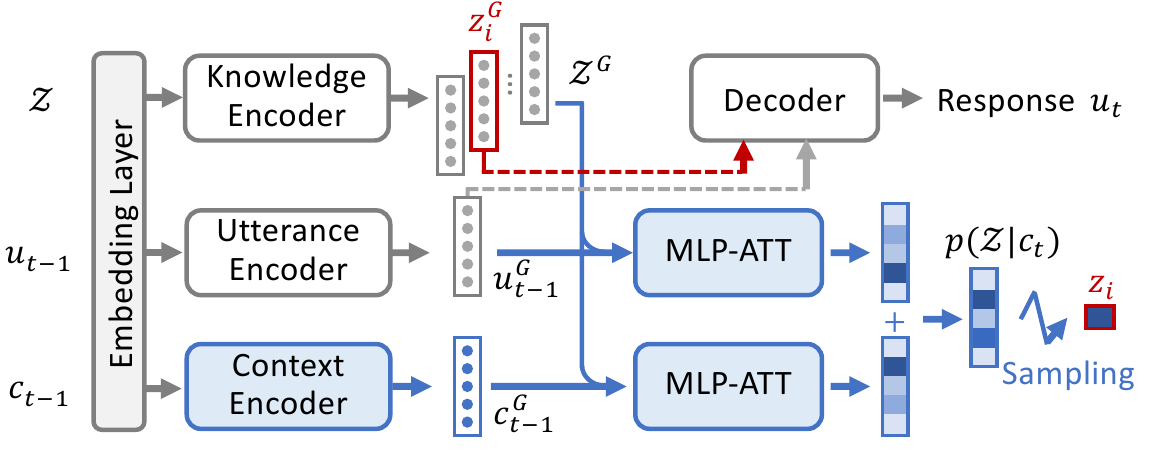}
	\caption{Architecture of dialogue generation.}
	\label{fig:Infrastructure}
\end{figure}

The detailed network architecture of dialogue generation is illustrated in Figure \ref{fig:Infrastructure}. With the context and background knowledge as input, our dialogue strategy selects one piece of appropriate knowledge to generate informative and coherent response. The background $\mathcal{Z}=\{z_1,z_2,\cdots,z_M\}$ includes a set of knowledge, where a piece of knowledge $z_i$ is presented by one sentence, such as ``i like to ski". Utterance $u_{t-1}$ is the last response from the other participant and the context $c_t=\text{concat}(u_1, u_2, \cdots, u_{t-1})$ is the current conversation history.

It is worth noting that in our dialogue generation, the input context $c_t$ is separated into two parts, with independent encoders employed for utterance $u_{t-1}$ and context $c_{t-1}$ respectively. The motivation to do so lies in two aspects: for the sake of coherence, the knowledge utilized in $t$-th turn is supposed to be semantically related to the partner's last utterance $u_{t-1}$; to avoid repetition, the knowledge utilized in $t$-th turn should be dissimilar with the former dialogue history $c_{t-1}$. 

After passing through the embedding layer and the encoders of gated recurrent unit (GRU) \cite{cho2014properties}, the inputs obtain their corresponding feature representation: knowledge $z_i^G$, utterance $u_{t-1}^G$ and context $c_{t-1}^G$. $\mathcal{Z}^G=\{z_1^G,z_2^G,\cdots,z_M^G\}$ is the set of knowledge representation. With discriminative representations $u_{t-1}^G$, $c_{t-1}^G$ and $\mathcal{Z}^G$ obtained, the prior distribution over knowledge $p(\mathcal{Z}|c_t)$ can be estimated through MLP attention (MLP-ATT) \cite{bahdanau2014neural}:
\begin{equation}\label{eq:att}
	\begin{split}
		&p(\mathcal{Z}|c_t)=p(\mathcal{Z}|u_{t-1})*0.5+p(\mathcal{Z}|c_{t-1})*0.5,\\
		&p(z_i|u_{t-1})=\text{softmax}\big(\text{MLP-ATT}(u_{t-1}^G, z_i^G)\big),~~~~~~\\
		&p(z_i|c_{t-1})=\text{softmax}\big(\text{MLP-ATT}(c_{t-1}^G, z_i^G)\big),
		\raisetag{1.9\baselineskip}
	\end{split}
\end{equation}
where softmax is defined as $\text{softmax}(s_i)=e^{s_i} /\sum_j e^{s_j}$ \cite{sukhbaatar2015end}. 
And the computation of MLP-ATT is given as follows:
\begin{equation}\nonumber
\text{MLP-ATT}(x,y)=V_1^T \tanh(x W_1 + y W_2),
\end{equation}
where $W_1, W_2 \in \mathbb{R}^{d\times d}$ and $V_1\in \mathbb{R}^d$ are the weight matrices. $p(\mathcal{Z}|c_t)$ is the probability distribution for knowledge selection and $\sum_{i=1}^M p(z_i|c_t) =1$. (If $p(z_i|c_t)=0.2$, it means that the probability to select knowledge $z_i$ is $0.2$.) According to the estimated prior probability distribution $p(\mathcal{Z}|c_t)$, one piece of knowledge can be sampled $z_i\sim p(\mathcal{Z}|c_t)$ and sent to the decoder for response generation $p(u_t|z_i,u_{t-1})$. 

It is obvious that the key component for informative and coherent conversation is the appropriate knowledge selection, shown as Blue areas in Figure \ref{fig:Infrastructure}. Nevertheless, a high-fidelity decoder $p(u_t|z_i,u_{t-1})$, which is able to express the given knowledge accurately, is also indispensable. To this end, the pre-training is carried out using those target responses associated with ground-truth knowledge via supervised learning. The training data is in the format of $\{u_{t-1}, ~z_i, ~u_t\}$,  where $u_{t-1}$ is the last utterance from the partner, $u_t$ is the target response and $z_i$ is the ground truth knowledge used in $u_t$. Major steps in the pre-training are listed as follows: (1) the encoders convert the knowledge and utterance into $z_i^G$ and $u_{t-1}^G$; (2) the decoder tries to generate the response $u_t$ based on the ground-truth knowledge $z_i$ and last utterance $u_{t-1}$; (3) parameters in the encoders and decoder (Gray areas) are optimized via supervised leaning, with the loss functions defined in \newcite{zhao2017learning}. For the rest of the parameters related to the knowledge selection strategy (Blue areas), they will keep evolving through Generation-Evaluation reinforcement learning, which will be discussed in detail. 

\subsection{Strategy Evaluation}
Multi-turn knowledge grounded conversations are generated by two dialogue agents. To evaluate the effectiveness of deployed strategy, generated conversations and agents' background knowledge are collected for evaluation and two metrics are judiciously designed -- informativeness and coherence.

\subsubsection{Informativeness}
Information is a crucial ingredient in generating meaningful conversations. Although many approaches have been introduced to boost the generation of informative utterances, due to a lack of thorough control on effective information utilization, they are prone to generating repetitive utterances in multi-turn conversations. In this paper, we design a novel informativeness metric to measure the effective exploitation of information in the conversation level, which encourages extensive coverage and avoids unnecessary repetition. 
\begin{figure}
	\centering
	\includegraphics[width=0.48\textwidth]{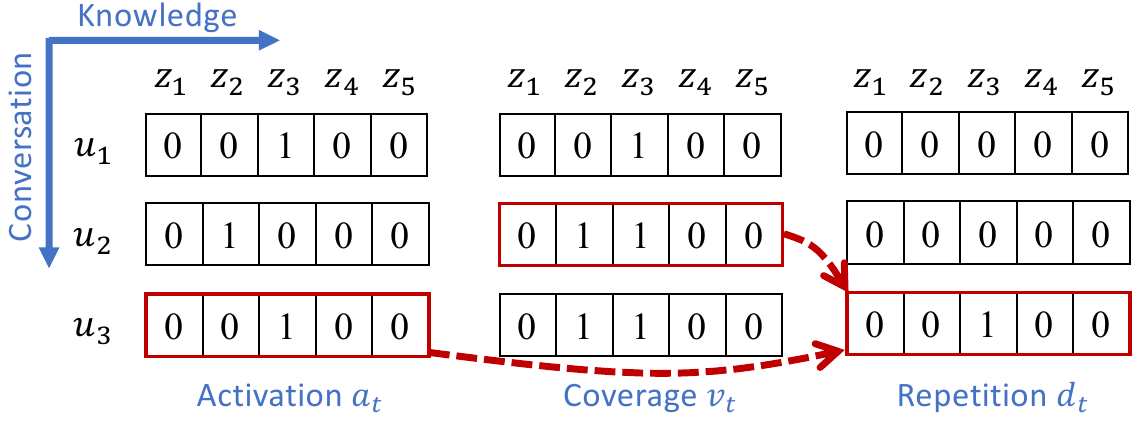}
	\caption{Toy example of informativeness assessment: activation $a_t$ records whether a piece of knowledge is expressed in $u_t$, coverage $v_t$ keeps track of expressed knowledge and repetition $d_t$ detects reiteration.}
	\label{fig:Coverage-Repeat}
\end{figure}

To illustrate the informativeness assessment, a toy example is given in Figure \ref{fig:Coverage-Repeat}. Assume that there are five pieces of background knowledge $z_i$ within the conversation participants. For each generated utterance $u_t$, it will be assessed whether $z_i$ is expressed by $u_t$ or not, which can be approximately inferred through keyword matching (in the form of binary variable 0/1). Such estimation over the background knowledge is stored in the activation vector $a_t$. If relying on $a_t$ as the informativeness metric, it is able to boost informative response generation on the utterance level. However, it inevitably produces repetitive responses due to the lack of information utilization control on the conversation level.

Inspired by the coverage mechanism in machine translation \cite{tu2016modeling} and text summarization \cite{see2017get}, we propose to maintain one coverage vector $v_t$ to keep track of the activation on each piece of information during the conversation flow. From the toy example, it can be observed that the coverage vector $v_t$ increases with the amount of expressed knowledge. In other words, a higher mean value of $v_t$ indicates that the participants have expressed more background knowledge, which gives a better chance for them to know more about each other.

Although the coverage mechanism stimulates extensive knowledge expression, it still lacks effective and explicit control on the reiteration. For the sake of user experience, we also maintain one repetition vector $d_t$ to detect information redundancy, whose estimation is carried out by jointly considering current information activation and last-step coverage status:
\begin{equation}
	d_t=\min(a_t, v_{t-1}),
\end{equation}
where the function $\min (\cdot)$ calculates the element-wise minimum value between two vectors. As shown in Figure \ref{fig:Coverage-Repeat}, when utterance $u_3$ reiterates the same information as before, it does not increase knowledge coverage and leads to unnecessary repetition. 

In summary, instead of focusing on the information activation of the single-round response, our informativeness metric considers the effective information utilization in the scope of multi-turn conversation. For a conversation with $T$ turns, its informativeness is estimated as follows:
\begin{equation}
	r_I = \text{mean}(v_T) - \sum_{t=1}^T \text{mean}(d_t),
\end{equation}
where the function $\text{mean}(\cdot)$ calculates the mean value of a vector. By maintaining information coverage and internal repetition simultaneously, the conversation level informativeness is able to encourage informative and concise conversations.

\subsubsection{Coherence}
For the sake of natural interaction, coherence is another indispensable ingredient in strategy evaluation. In addition to relevance with the context, the coherence assessment also evaluates the conversation consistency with the backgrounds. The motivation to enforce background consistency is to confine the massive and loose interactive responses into a reasonable space. Considering that the essence of coherence is semantic relevance between two inputs and many deep learning based approaches have demonstrated their superiority at capturing semantic relevance, such as DSSM \cite{huang2013learning}, SMN \cite{wu2017sequential} and BERT \cite{devlin2018bert}, we use a symmetric neural network for the coherence assessment in this paper.
\begin{figure}
	\centering
	\includegraphics[width=0.48\textwidth]{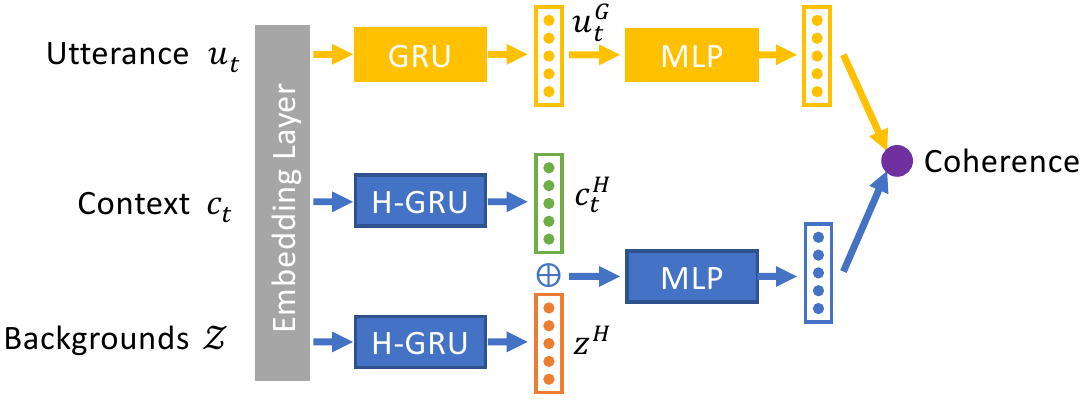}
	\caption{Illustration of coherence assessment, where H-GRU refers to hierarchical GRU and the symbol \textcolor{blue}{$\oplus$} denotes vector concatenation.}
	\label{fig:Coherence}
\end{figure} 

As shown in Figure \ref{fig:Coherence}, for a generated utterance $u_t$, its coherence with the context $c_t$ and corresponding backgrounds $\mathcal{Z}$ can be estimated through this symmetric network. The utterance is fed into the embedding layer, followed by gated recurrent unit (GRU) \cite{cho2014properties} and multilayer perceptron (MLP) to capture discriminative representation. As for the context and backgrounds, they are fed into the embedding layer and the hierarchical GRU for better feature extractions \cite{sordoni2015hierarchical}, which are further concatenated together to obtain comprehensive representation. The final coherence is estimated as the inner product between two vectors:
\begin{equation}\label{eq:Coherence}
\begin{split}
&r_{C_t} = \sigma\big(\text{MLP}(u_t^G) \cdot \text{MLP}([c_t^H,z^H]) \big),\\
&\text{where}~~\text{MLP}(x)=\sigma(x W_1+b_1) W_2 +b_2~.
\end{split}
\end{equation}
$\sigma(\cdot)$ is the sigmoid activation, $[\cdot,\cdot]$ denotes vector concatenation and MLP includes two linear transformations with a sigmoid activation in between. 

The above equation evaluates the coherence for each generated utterance $u_t$, by considering existing conversation history and corresponding background, which is further summed up over all utterances as conversation-level coherence assessment. 

\subsubsection{Compound Assessment}
To provide a united reinforcement signal for strategy evolution, the informativeness and coherence assessments are further integrated as a compound reward. For a conversation $\tau$ with $T$ turns, the compound assessment is defined as:
\begin{equation}
	R(\tau)=\sum_{t=1}^T r_{C_t} + r_I~.
\end{equation} 
The two intrinsic factors in human conversations -- exchanging information and enhancing interaction have been included in our compound reward. 

\subsection{Strategy Evolution}\label{sec:evolve}
From the perspective of reinforcement learning, the knowledge selection within a conversation can be regarded as sequential actions taken within a trajectory. As such, the objective of knowledge grounded dialogue generation can be written as:
\begin{equation}
	\max ~J(\theta) = \mathbb{E}_{\tau \sim p(\tau;\theta)}R(\tau),
\end{equation}
where  $\theta$ refers to the network parameters of dialogue generation, $\tau \sim p(\tau;\theta)$ is a multi-turn conversation generated under the deployed strategy and $R(\tau)$ is the compound assessment of strategy evaluation. Gradient update of the above objective can be further derived as follows:
\begin{equation}\label{eq:RL-full}
\small
\begin{split}
\triangledown_\theta J(\theta) 
= &\sum_{t=1}^T \triangledown_\theta \log \big(p(z_i|c_t)p(u_t|z_i,u_{t-1})\big)\big(R(\tau)-b\big) ,\\
=&\sum_{t=1}^T \triangledown_\theta \log p(z_i|c_t)\big(R(\tau)-b\big)\\
&+\sum_{t=1}^T \triangledown_\theta \log p(u_t|z_i,u_{t-1})\big(R(\tau)-b\big),
\raisetag{3.8\baselineskip}
\end{split}
\end{equation}
where $b$ is the reward baseline estimated with $K$ times Monte Carlo sampling: $b=\sum_k R(\tau^{(k)})/K$. In Equation \eqref{eq:RL-full}, the first term is about the dialogue strategy of appropriate knowledge selection and the second term is about the decoding process with the selected knowledge. As suggested in \citep{lewis2017deal, Yarats2018HierarchicalTG}, applying reinforcement learning on the decoder might lead to poor linguistic quality. As such, in this paper, the focus is on the strategy evolution and gradient update is further simplified:
\begin{equation}\label{eq:RL-partgrad}
	\triangledown_\theta J(\theta) = \sum_{t=1}^T \triangledown_\theta \log p(z_i|c_t)\big(R(\tau)-b\big)~.
\end{equation} 
The physical meaning of the above equation is given as follows: the strategies that lead to higher conversation rewards will be encouraged and those that result in lower conversation rewards will be suppressed. 

As demonstrated in Equation \eqref{eq:RL-partgrad}, the network parameters related to dialogue strategy (Blue areas in Figure \ref{fig:Infrastructure}) will keep evolving via compound assessment. For the rest parameters, they are pre-trained with supervised learning and will be kept fixed during strategy evolution.

\section{Experiments}
\subsection{Settings}
All experiments have been carried out on the publicly available dataset -- PersonaChat \cite{zhang2018personalizing}, which provides both human annotated conversations and the participants' background knowledge (persona profiles). PersonaChat has separated training and testing set. In total, there are 8,939 dialogues (131,438 turns) in the training set and 968 dialogues (15,024 turns) in the testing set. 
Comprehensive comparisons have been made to the following methods:
\begin{itemize}
\item Sequence to sequence with attention (Seq2Seq) \cite{vinyals2015neural} is the classic response generation approach, without using any extra knowledge.

\item The knowledge grounded memory network (Mem-Net) \cite{ghazvininejad2018knowledge} encodes text knowledge into memory to boost the generation of informative responses. 

\item The KG-Net \cite{lian2019learning} makes use of posterior knowledge distribution in the training process for accurate informative response generation and achieves the state-of-the-art results on PersonaChat.

\item \newcite{li2016deep} first employed reinforcement learning for dialogue generation (RL-DG), where simple Seq2Seq was used as the generation model. In the experiments, to improve RL-DG's performance, KG-Net is utilized as the base model for informative generation.
\end{itemize}

In our strategic knowledge interaction, the parameters of knowledge encoder, utterance encoder and decoder were pre-trained with supervised learning. For the learnable parameters (Blue areas in Figure \ref{fig:Infrastructure}), the context encoder was initialized with the utterance encoder and random initialization was employed for the rest layers\footnote{Our code and model will be released at \url{https://github.com/PaddlePaddle/models/tree/develop/PaddleNLP/Research/ACL2019-SEEDS}.}. The training process was carried out using Adam optimizer, with a learning rate of 2e-4. The conversation turns $T$ was set to 8, batch size was set to 8 and Monte Carlo sampling times $K$ was set to 16. 

\subsection{Experimental Results}
\begin{figure*}
	\centering
	\includegraphics[width=\textwidth]{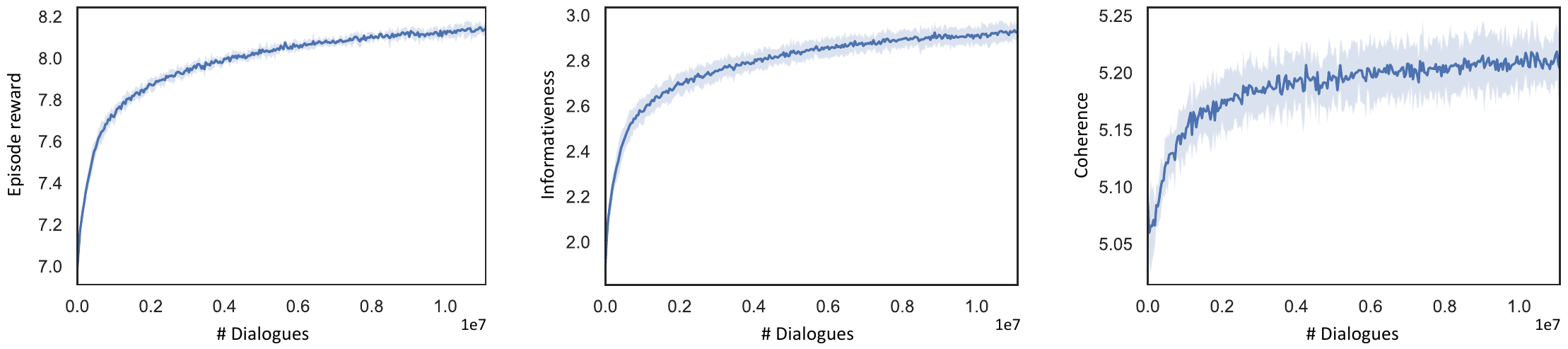}
	\caption{Training curves of reinforcement learning.}
	\label{fig:curve}
\end{figure*}

The training curves of reinforcement learning are shown in Figure \ref{fig:curve}, which are the results averaged over 5 random seeds. The horizontal axis refers to the number of trained dialogues. The vertical axis stands for the compound episode reward, informativeness and coherence, respectively. These results demonstrate that all rewards increase stably within the training process and remarkable increments are achieved after convergence. 

\subsubsection{Automatic Evaluation}
The experimental results with automatic measurements are summarized in Table \ref{tab:auto}, with highest value written in bold. Distinct-1/2 \cite{li2015diversity} measures the diversity of generated conversations, which is defined as the amount of distinct unigrams or bigrams divided by the total number of generated words. Knowledge-Recall/Precision/F1 \cite{dinan2018wizard} measures the informativeness of generated conversations with regarding to background knowledge, defined as:
\begin{equation}
\begin{aligned}
&\text{Recall}=\frac{|W_G\cap W_K|}{|W_K|}~,\\
&\text{Precision}=\frac{|W_G\cap W_K|}{|W_G|}~,\\
&\text{F1}=2\times \frac{\text{Recall}\times \text{Precision}}{\text{Recall}+\text{Precision}}~,
\end{aligned}
\end{equation}
where $W_G$ and $W_K$ refer to the set of non-stop words in generated conversations and background knowledge. 

\begin{table}
	\centering
	\includegraphics[width=0.48\textwidth]{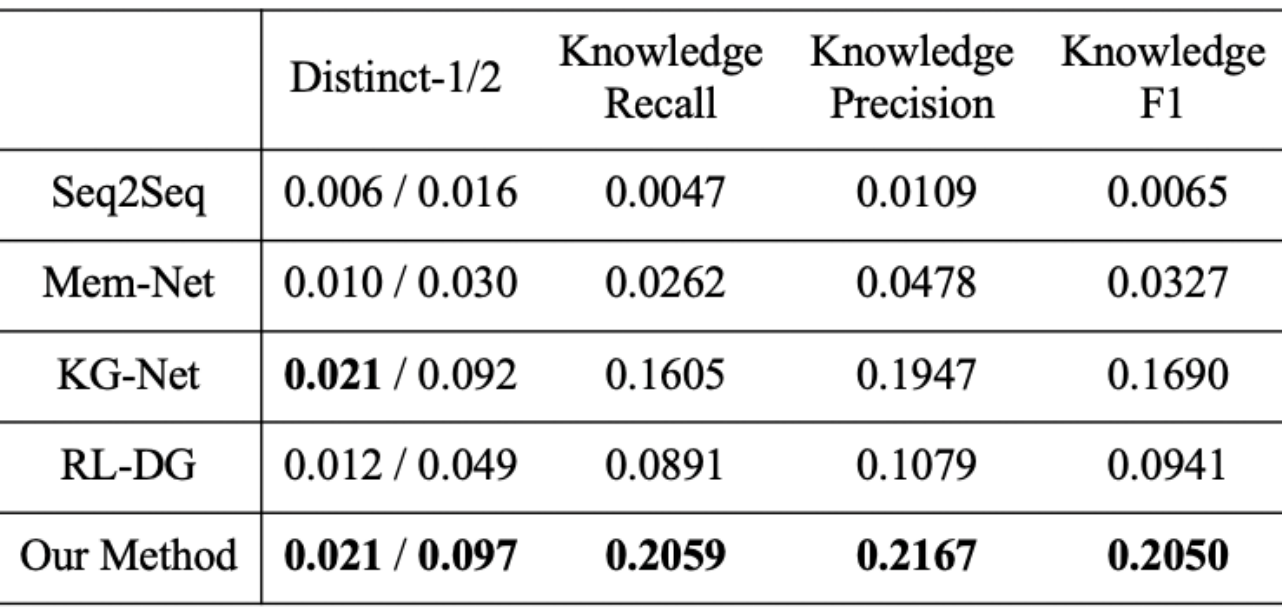}
	\caption{Experimental results with automatic measurements, with highest value written in bold.}
	\label{tab:auto}
\end{table} 
From Table \ref{tab:auto}, it demonstrates that the proposed method obtains the best results. The distinct measurement indicates that more diverse words or phrases are produced by our method. The knowledge measurement verifies the effectiveness of our approaches on the knowledge utilization in multi-turn conversations. As compared with the state-of-the-art KG-Net, the knowledge F1 of our method is increased by 3.6\%, which is a significant improvement. 

\subsubsection{Human Evaluation}
Currently, most automatic metrics are not aligned well with human beings in dialogue evaluation \cite{liu2016not}, such as BLEU, ROUGE, etc. In our experiments, extensive evaluations have been carried out with crowd-sourced human beings. With the background knowledge (persona profiles of two participants) and the first start utterance in the testing set, simulated dialogues were generated using each method. There are 8 turns in the simulated conversations (1 start utterance followed by 7 successive generated responses). 

Our method is compared with the rest state-of-the-art approaches and each group contains 100 pairs of simulated dialogues, randomly selected from the testing set. For each pair of conversations, they share the same background knowledge and 3 crowd-sourced workers are asked to compare these two simulated conversations at the same time. The human evaluations include the following aspects: (1) Overall refers to the general preference towards the two conversations, with a joint consideration of effective information exchange and coherent interaction. (2) Coverage measures the amount of knowledge expressed during conversations. (3) Concise considers the information repetition and utterance reiteration within conversations. (4) Coherence estimates the consistency and appropriateness within the interaction between participants. 

The final comparison results by crowd-sourced workers are determined through majority voting, which are summarized in Table \ref{tab:human}. These results demonstrate that our method is consistently and significantly better than the other state-of-the-art approaches. 

\begin{table*}
	\centering
	\includegraphics[width=\textwidth]{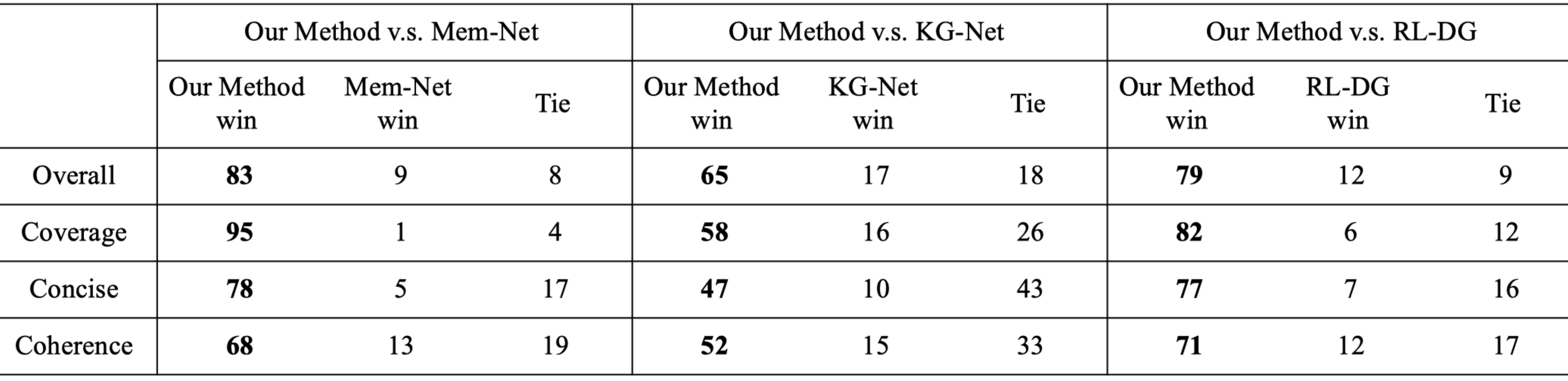}
	\caption{Experimental results with human evaluation, with highest value written in bold.}
	\label{tab:human}
\end{table*} 
\begin{table*}
	\centering
	\includegraphics[width=\textwidth]{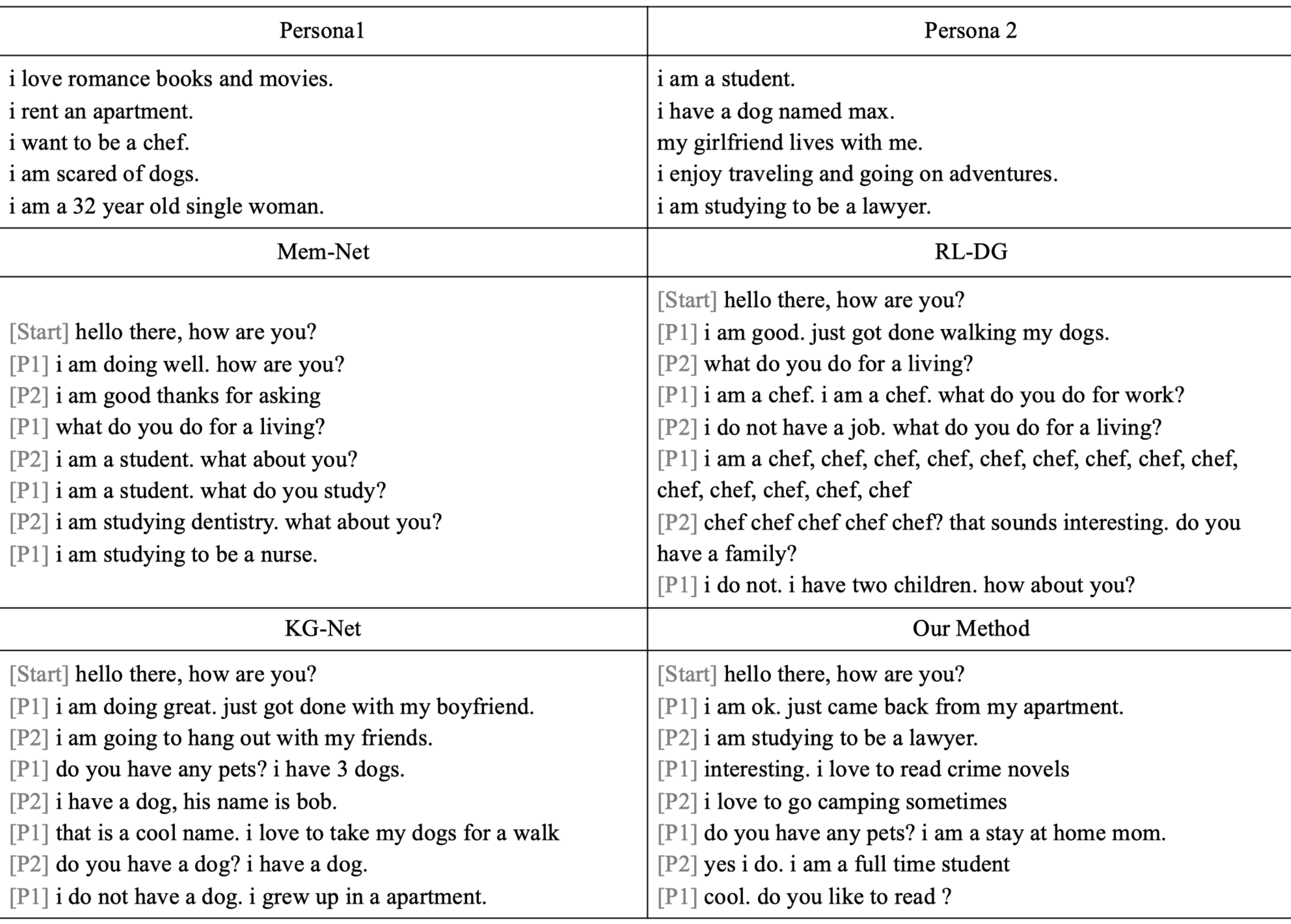}
	\caption{Simulated dialogues with the same personas and start utterance.}
	\label{tab:case}
\end{table*} 
\begin{figure*}
	\centering
	\includegraphics[width=\textwidth]{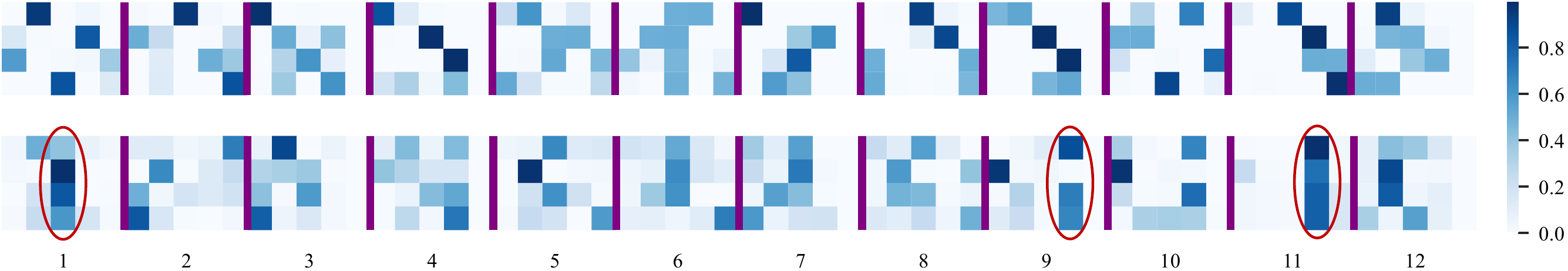}
	\caption{Visualisation of knowledge utilization in conversations of our method (Upper) and KG-Net (Bottom). Horizontal: background knowledge in the first 12 simulated dialogues, separated by Purple lines. Vertical: knowledge selection probability of each response by one participant. }
	\label{fig:visual}
\end{figure*} 
\subsection{Discussions}
\subsubsection{Case Analysis}
Table \ref{tab:case} provides several detailed cases of the simulated dialogues generated by each method, under the same background knowledge (persona profiles) and the start utterance. It can be observed that Mem-Net tends to generate general and fluent responses, like ``what about you", while expresses limited background knowledge. Although informative utterances can be generated by KG-Net, due to a lack of control on information utilization, serious repetition has emerged in the simulated conversation. In addition to redundant responses, another problem with RL-DG is the poor linguistic quality, which might be caused by the decoder update via RL \citep{lewis2017deal, Yarats2018HierarchicalTG}. Our method is able to generate informative and coherent conversations because the decoder is fixed and only the knowledge selection strategy keeps evolving via compound assessment

Visualization of knowledge utilization in conversations is displayed in Figure \ref{fig:visual}, where the first 12 simulated dialogues from the testing set are presented. The horizontal axis is the background knowledge in the dialogues, separated by Purple lines. The vertical axis shows the knowledge selection probability $p(z_i|c_t)$ of each utterance, made by one participant in the simulated dialogues (in total 4 utterances). The upper part (our method) demonstrates extensive knowledge coverage, while the bottom part (KG-Net) exhibits repetitive knowledge utilization (highlighted with red circles). 

\subsubsection{Correlation Analysis} 
The correlation statistics between automatic metrics (including the distinct-1/2, knowledge-R/P/F1 and our compound reward) and human annotations are provided in Table \ref{tab:corr}. The Pearson correlation coefficient \cite{benesty2009pearson} is estimated using the annotated overall score of our method v.s. KG-Net. These results indicate our designed compound reward is aligned better with human beings than commonly used metrics.
\begin{table}
	\centering
	\includegraphics[width=0.48\textwidth]{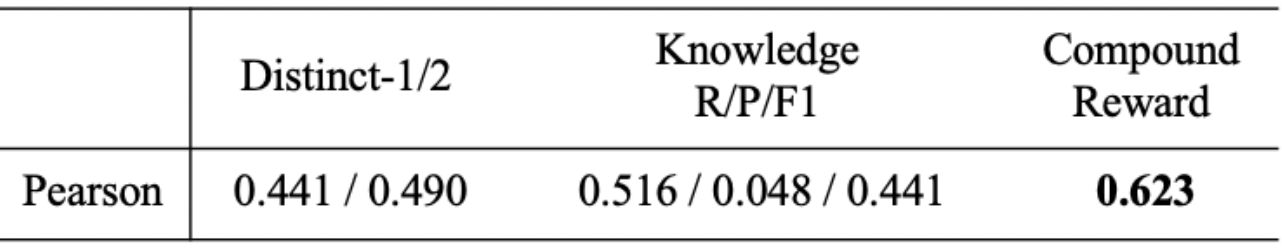}
	\caption{Correlation between automatic metrics and human evaluations, with highest value written in bold.}
	\label{tab:corr}
\end{table}

\subsubsection{Further Evaluation of the Dialogue Strategy}
The PersonaChat dataset is also employed by the ConvAI2 challenge \cite{dinan2019second}, where the team Lost in Conversation obtained the best performance. The network of Lost in Conversation involves 12 transformer layers, which requires extra training data in addition to PersonaChat. For fair comparison, our dialogue strategy is also implemented with the same number of transformer layers and training settings used by Lost in Conversation. The comparison is summarized in Table \ref{tab:LIC}, which verifies the superiority of our proposed method over the advanced transformer network. 
\begin{table}
	\centering
	\includegraphics[width=0.48\textwidth]{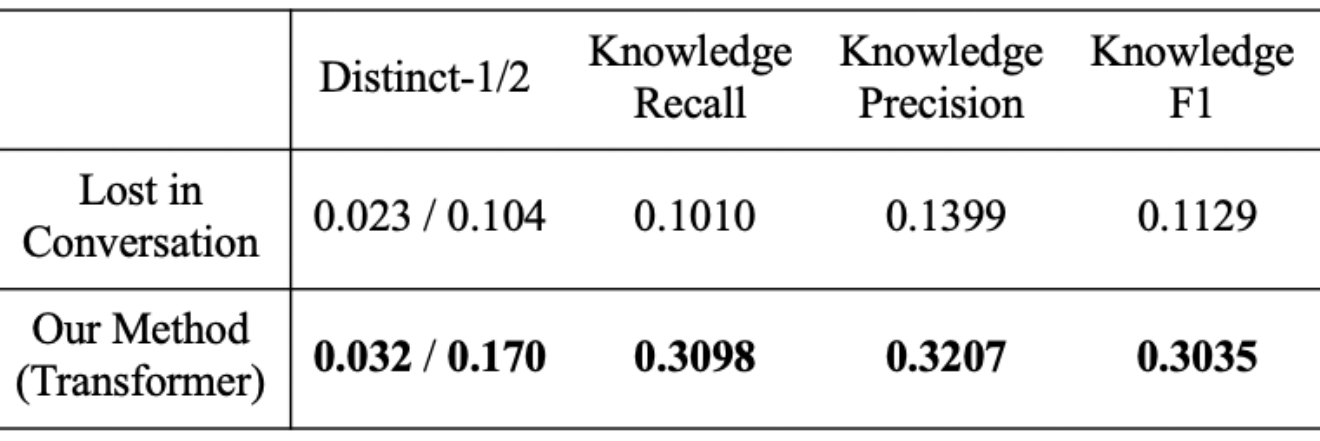}
	\caption{Comparison with Lost in Conversation, with highest value written in bold.}
	\label{tab:LIC}
\end{table} 

\section{Related Work}
Our work is related with knowledge grounded response generation and multi-turn conversation with reinforcement learning.

As conventional Seq2Seq \cite{vinyals2015neural} tends to generate general and dull responses, some knowledge grounded approaches have been introduced to increase the informativeness with extra knowledge. MemNet \cite{ghazvininejad2018knowledge} encodes factual texts into memory and decodes via attention mechanism for informative generation. CCM \cite{zhou2018commonsense} relies on structured knowledge to generate rich-information response. In \newcite{lian2019learning}, the posterior distribution is estimated and accurate knowledge is selected to boost informative generation. However, without thorough consideration and control on the knowledge utilization in multi-turn conversations, the above approaches are prone to produce repetitive and incoherent utterances.

The technique of reinforcement learning has been applied to multi-turn dialogue systems in several scenarios. In RL-DG \cite{li2016deep}, three rewards are defined and combined together to boost diverse response generation. Due to a lack of effective control on knowledge utilization, RL-DG is unable to express extensive information during conversations. As RL-DG relies on the reinforcement signal to update all components in the dialogue system, including decoder, it suffers from poor linguistic quality. In \newcite{yao2018chat}, reinforcement learning is employed to plan a cue word (topic) path for a dialogue, where the cue word at $t$-th turn will assist the corresponding response generation. Different from these chit-chat approaches, our dialogue generation is conducted under the objective of facilitating effective information exchange and letting both participates know more about each. With judiciously design of evaluation metrics, our compound reward is aligned well with human beings and provides meaningful reinforcement signal to evolve the dialogue strategy.

\section{Conclusion}
In this paper, a novel Generation-Evaluation framework is proposed for informative and coherent multi-turn dialogue generation. Knowledge grounded conversations are generated under the dialogue strategy, which is able to continuously evolve via reinforcement learning with the compound reward. Comprehensive experimental results demonstrate that the proposed method obtains superior performances than the other state-of-the-art methods on both automatic measurements and human evaluations. 

In the future, our work can be potentially improved by enriching the assessments with more fine-grained criteria, which can fully integrate turn-level cohesion and dialogue-level coherence. We will also explore to make full use of  knowledge to guide the selection of policy strategies for multi-turn conversation.

\section*{Acknowledgments}
We would like to thank the ACL reviewers for their constructive suggestions and Jinhua Peng, Chaotao Chen, Min Xie for the helpful discussions. This work was supported by the Natural Science Foundation of China (No.61533018).

\bibliography{Strategic}
\bibliographystyle{acl_natbib}

\end{document}